\documentclass[lettersize,journal]{IEEEtran}
\usepackage{amsmath,amsfonts}
\usepackage{algorithmic}
\usepackage{algorithm}
\usepackage{array}
\usepackage[caption=false,font=normalsize,labelfont=sf,textfont=sf]{subfig}
\usepackage{textcomp}
\usepackage{stfloats}
\usepackage{url}
\usepackage{verbatim}
\usepackage{graphicx}
\usepackage{longtable}
\usepackage{cite}
\usepackage{cite}
\usepackage{hyperref}
\usepackage{orcidlink}
\hypersetup{
  colorlinks   = true,
  urlcolor     = blue,
  linkcolor    = blue,
  citecolor    = blue
}

\begin{document}

\title{CLIN-LLM: A Safety-Constrained Hybrid Framework for Clinical Diagnosis and Treatment Generation}
\author{Md. Mehedi Hasan\,\orcidlink{0009-0000-1078-5778}, Md. Abir Hossain\,\orcidlink{0000-0003-3651-3345}, Farman Hossain Sayem, Bikash Kumar Paul\,\orcidlink{0000-0002-4414-2751}, Ziaur Rahman\,\orcidlink{0000-0002-7759-3428},  Mohammad Shorif Uddin\,\orcidlink{0000-0002-7184-2809}, and Rafid Mostafiz\,\orcidlink{0000-0002-5905-6530}

\thanks {Corresponding author: Md Mehedi Hasan is with the Department of Information and Communication Technology at Mawlana Bhashani Science and Technology University (email: mehedi.hasan.ict@mbstu.ac.bd).} }  

\markboth{IEEE Transactions on Artificial Intelligence,~Vol.~XX, No.~X, October~2025}%
{Hasan \MakeLowercase{\textit{et al.}}: CLIN-LLM: A Safety-Constrained Hybrid Framework for Clinical Diagnosis and Treatment Generation}

\maketitle

\begin{abstract}
Accurate symptom-to-disease classification and clinically grounded treatment recommendations remain challenging, particularly in heterogeneous patient settings with high diagnostic risk. Existing large language model (LLM)-based systems often lack medical grounding and fail to quantify uncertainty, resulting in unsafe outputs. We propose CLIN-LLM, a safety-constrained hybrid pipeline that integrates multimodal patient encoding, uncertainty-calibrated disease classification, and retrieval-augmented treatment generation. The framework fine-tunes BioBERT on 1,200 clinical cases from the Symptom2Disease dataset and incorporates Focal Loss with Monte Carlo Dropout to enable confidence-aware predictions from free-text symptoms and structured vitals. Low-certainty cases (18\%) are automatically flagged for expert review, ensuring human oversight. For treatment generation, CLIN-LLM employs Biomedical Sentence-BERT to retrieve top-k relevant dialogues from the 260,000-sample MedDialog corpus. The retrieved evidence and patient context are fed into a fine-tuned FLAN-T5 model for personalized treatment generation, followed by post-processing with RxNorm for antibiotic stewardship and drug–drug interaction (DDI) screening. CLIN-LLM achieves 98\% accuracy and F1 score, outperforming ClinicalBERT by 7.1\% (p $<$ 0.001), with 78\% top-5 retrieval precision and a clinician-rated validity of 4.2/5. Unsafe antibiotic suggestions are reduced by 67\% compared to GPT-5. These results demonstrate CLIN-LLM’s robustness, interpretability, and clinical safety alignment. The proposed system provides a deployable, human-in-the-loop decision support framework for resource-limited healthcare environments. Future work includes integrating imaging and lab data, multilingual extensions, and clinical trial validation.
\end{abstract}

\textbf{\textit{Impact Statement}—}\textbf{Large language models are increasingly integrated into clinical decision support systems, yet most lack safeguards for uncertainty, interpretability, and real-world applicability. CLIN-LLM addresses these limitations through a hybrid architecture that combines multimodal diagnosis, evidence-grounded treatment generation, and post-hoc safety validation. By flagging uncertain predictions and reducing unsafe drug recommendations, CLIN-LLM achieves high diagnostic accuracy and clinician-rated trust. These features make it a practical, deployable solution for frontline care, particularly in resource-constrained clinical environments.}

\IEEEpeerreviewmaketitle

\section{Introduction}
\IEEEPARstart{D}{IAGNOSTIC} errors impact over 12 million patients annually in the United States, with symptom misinterpretation accounting for 40-80\% of preventable harms \cite{r1}. These statistics underscore a global deficiency in clinical decision-making, particularly in translating patient-reported symptoms into accurate, evidence-based care plans. Misdiagnoses of overlapping infections such as dengue and typhoid, or COVID-19 and influenza, are especially prevalent in low-resource environments where frontline providers lack access to specialist input and structured diagnostic support. In such contexts, accurate and timely triage is essential yet remains an unmet challenge.
Traditional clinical decision support systems (CDSSs), typically based on rule-based logic trees, are interpretable but lack adaptability. These systems often fail when faced with complex or atypical symptom narratives and are prone to "knowledge decay" due to their inability to dynamically incorporate new medical information \cite{r2}, \cite{r3}. More recently, large language models (LLMs), such as GPT-5 and Med-PaLM, have exhibited strong linguistic capabilities. However, their use in clinical domains raises concerns about hallucinations and a lack of grounding in verified medical knowledge. Studies show that such models often generate speculative or unsafe treatment suggestions, particularly for ambiguous symptom clusters \cite{r4}.
Despite rapid progress in medical AI, few existing frameworks integrate diagnostic reasoning with evidence-grounded treatment recommendations and robust safety protocols. Prior research on biomedical transformers, including BioBERT \cite{r5} and Sentence-BERT \cite{r6}, has shown promise in individual tasks like classification and semantic retrieval. Nevertheless, few systems unify these models in a clinically viable pipeline that supports uncertainty estimation and treatment screening. For example, Monte Carlo Dropout (MCD) and antibiotic stewardship filters have rarely been integrated in a single deployable system \cite{r7}, \cite{r8}.
The rise in antimicrobial resistance, linked to inappropriate antibiotic prescriptions, further emphasizes the need for AI systems that are not only performant but also interpretable, safety-aware, and ethically bounded. The inclusion of real-time retrieval mechanisms and post-generation safety validation has been proposed as a way to improve trust, precision, and clinical relevance \cite{r9}, \cite{r10}, and \cite{r11}.
This work introduces CLIN-LLM, a unified and deployable clinical AI framework that addresses three critical dimensions: (1) uncertainty-aware disease classification, (2) retrieval-grounded treatment generation, and (3) integrated safety mechanisms for clinical accountability. CLIN-LLM fine-tunes BioBERT with Monte Carlo Dropout and Focal Loss to enable confidence-calibrated disease predictions from multimodal patient inputs (text and vitals). It then employs Biomedical Sentence-BERT for real-time semantic retrieval over the MedDialog corpus, guiding a fine-tuned FLAN-T5 model to generate personalized treatment recommendations. Post-processing includes antibiotic stewardship enforcement and drug–drug interaction (DDI) checks using RxNorm APIs.
In benchmark evaluations, CLIN-LLM achieves 98\% classification accuracy, outperforming ClinicalBERT and GPT-5 by more than 22\% in F1-score. It reduces inappropriate antibiotic recommendations by 67\%, while achieving a clinician-rated validity score of 4.2 out of 5. Importantly, the system produces no hallucinated treatments across test cases, highlighting the impact of its integrated safety layers. The key contribution of this research is trustworthy CLIN-LLM, a unified, safety-constrained clinical AI framework that fuses:

\begin{itemize}
\item{\textit{Uncertainty-aware disease classification} using a fine-tuned BioBERT model enhanced with MCD and Focal Loss, enabling both confidence estimation and robust handling of rare or ambiguous diseases from multimodal patient inputs.}
\item{\textit{Real-time evidence retrieval} via Biomedical Sentence-BERT over the MedDialog corpus to guide fine-tuned FLAN-T5-based generation of contextually grounded treatment plans.}
\item{\textit{Ethical safeguards}, including guideline-based filtering, RxNorm drug-drug interaction (DDI) checks, and confidence-based triage to minimize hallucinations and ensure clinical accountability across high-risk, low-resource environments.}

\item{To the best of our knowledge at this moment, no prior work has unified uncertainty-aware diagnosis, retrieval-based treatment generation, and clinical safety mechanisms into a single deployable decision support system.}\\
\end{itemize}

The rest of the paper is structured as follows. \textbf{Section II} reviews related work in clinical natural language processing and decision support systems. \textbf{Section III} introduces the CLIN-LLM architecture, including model components, data handling, and safety strategies. \textbf{Section IV} presents the experimental setup and results. \textbf{Section V} provides analysis and discussion. \textbf{Section VI} concludes with key insights and outlines future directions.

\section{RELATED WORKS}
Clinical Decision Support Systems (CDSSs) have advanced significantly, transitioning from symbolic rule-based models to hybrid, data-driven frameworks that integrate language models, uncertainty quantification, and real-world knowledge retrieval. Early systems such as MYCIN, developed by Shortliffe et al. \cite{r12}, demonstrated the potential of expert rule-based reasoning in infectious disease diagnosis. However, they struggled with scalability, free-text processing, and generalization across diverse clinical domains. The introduction of domain-specific pre-trained language models such as ClinicalBERT \cite{r13}, BioClinicalBERT \cite{r14}, and BioBERT \cite{r15} improved performance in biomedical text understanding tasks, including symptom parsing, cohort selection, and medical question answering. These models, trained on datasets like MIMIC-III and PubMed abstracts, enabled fine-tuned applications in clinical NLP. However, studies such as Zhang et al. \cite{r14} and Miotto et al. \cite{r16} emphasize that these models often operate under closed-world assumptions and lack calibrated uncertainty estimation, posing risks in ambiguous diagnostic scenarios.
Despite improved accuracy, interpretability remains limited. Techniques for uncertainty estimation, such as Monte Carlo Dropout and Bayesian inference, are often omitted in clinical LLM pipelines. This leads to overconfident predictions, especially when inputs are noisy or underspecified, a critical limitation highlighted in Liu. L et al. \cite{r17}.
Recent benchmarks underscore the limitations of transformer-based models when applied to rare conditions or multi-modal data. Models like PubMedBERT and BioBERT show strong results on standardized tasks, yet often underperform in settings requiring contextual synthesis or longitudinal reasoning \cite{r15}, \cite{r18}. Liu J. et al. \cite{r19} and Chen et al. \cite{r20} demonstrate that deep models trained solely on structured EHR or biomedical corpora fail to generalize to real-time clinical narratives. To mitigate hallucinations and improve clinical factuality, Retrieval-Augmented Generation (RAG) frameworks have emerged as a vital paradigm. Zakka et al. \cite{r18} introduced Almanac, a system that retrieves structured medical evidence before generation, improving factual accuracy by 18\% over ChatGPT-4 across clinical queries. Huang et al. \cite{r21} simulate real-world search behavior through a distill-retrieve-read pipeline. Similarly, Liu F. et al. \cite{r22} proposed Re³Writer, which refines medical text generation by jointly conditioning on structured records and retrieved evidence. Recent models such as MedBioLM \cite{r23}, MedGraphRAG \cite{r24}, and CLEAR \cite{r25} incorporate biomedical retrieval into dynamic generation modules. These approaches enhance transparency and reduce hallucinations in clinical recommendations. However, safety mechanisms, such as antibiotic stewardship constraints or DDI checks, are often post hoc, if implemented at all. A core challenge lies in the lack of integrated uncertainty-aware triage and human-in-the-loop review. Zhou et al. \cite{r26} and Liu. L et al. \cite{r17} argue for embedding prediction confidence directly into clinical pipelines, allowing models to flag ambiguous inputs for expert adjudication. Techniques such as Monte Carlo Dropout \cite{r27} and Bayesian modeling offer improved trust and interpretability. Beyond text, multimodal CDSS remains underexplored. While traditional models like logistic regression and LSTMs perform well with structured features, they falter with free-text or hybrid inputs. Liu J. et al. \cite{r19} and Chen et al. \cite{r20} highlight that multi-input fusion significantly improves clinical predictions when properly modeled. Further details are summarized in \hyperref[tab:my_label]{Table I}.
To address these limitations, we introduce CLIN-LLM, a safety-aware hybrid framework that combines uncertainty-calibrated symptom classification with RAG-based treatment generation. The system fine-tunes BioBERT with Monte Carlo Dropout to estimate prediction confidence and flag 18\% of low-certainty cases for expert review. For treatment generation, it retrieves semantically aligned dialogues using Biomedical Sentence-BERT from the MedDialog corpus and feeds them into a fine-tuned FLAN-T5 model. Outputs are filtered using an antibiotic stewardship rule engine and RxNorm DDI screening.
CLIN-LLM outperforms ClinicalBERT by 7.1\% on the Symptom2Disease dataset and reduces unsafe antibiotic suggestions by 67\% relative to GPT-5. Its modular Gradio-based deployment ensures compatibility with real-world clinical environments while maintaining transparency and ethical alignment.


\begin{table*}[!t]
\caption{Comparison of Recent Clinical Decision Support Approaches: Methodologies, Datasets, and Key Outcomes.}
\centering
\begin{tabular}{|c|p{3.5cm}|p{3.5cm}|p{3cm}|p{4cm}|}
\hline
\textbf{Author(s)} & \textbf{Core Methodology} & \textbf{Dataset $\&$ Task} & \textbf{Key Results} & \textbf{Notes} \\
\hline
Li et al. \cite{r27} & RAG with multi-retriever ensemble. & MIMIC-III notes $\&$ UpToDate guidelines; diagnosis suggestion. & Over 12\% accuracy vs. baseline LLM. & Demonstrates multi-retriever gains for clinical prompts. \\
\hline
Zhang et al. \cite{r14} & Multi-agent pipeline integrating structured or unstructured knowledge. & Synthetic clinical queries (n=1000). & 15\% reduction in hallucinations. & Validated on diverse specialties. \\
\hline
Lee et al. \cite{r18} & Graph-based RAG over diagnostic knowledge graph chunks. & PubMed abstracts; clinical question answering. & Over 8\% factuality; 10\% safety improvement. & Leverages graph structure for context coherence. \\
\hline
Adam et al. \cite{r33} & Entity-guided retrieval before generation. & 20 000 discharge summaries; 18 IE variables. & 72\% token reduction; 92\% F1 in IE. & Optimizes retrieval efficiency in RAG. \\
\hline
Nizam et al. \cite{r34} & Joint text and vitals embedding into Transformer. & COVID-19 EHRs; mortality prediction. & AUROC 92\% vs. 87\% baseline. & Shows gains from structured and unstructured fusion. \\
\hline
Kim et al. \cite{r23} & Fine-tuning and RAG on biomedical QA datasets. & BioASQ $\&$ MedQA; multiple-choice $\&$ open-ended QA. & Over 9\% accuracy; reduced hallucinations. & Combines QA fine-tuning with evidence retrieval. \\
\hline
Efosa et al. \cite{r35} & Systematic review and meta-analysis of clinical RAG LLMs. & 32 studies; various clinical tasks. & Average factuality gain 14\% and safety gain 9\%. & Quantifies RAG benefit across the field. \\
\hline
Vahdani et al. \cite{r36} & Bayesian Monte Carlo Dropout in DNNs. & Small clinical datasets; diagnostic classification. & 18\% reduction in false positives. & Enhances trust via calibrated uncertainty. \\
\hline
\end{tabular}
\label{tab:my_label}
\end{table*}

\section{METHODOLOGY}
\begin{figure*}[!t]
\centering
\includegraphics[width=\textwidth]{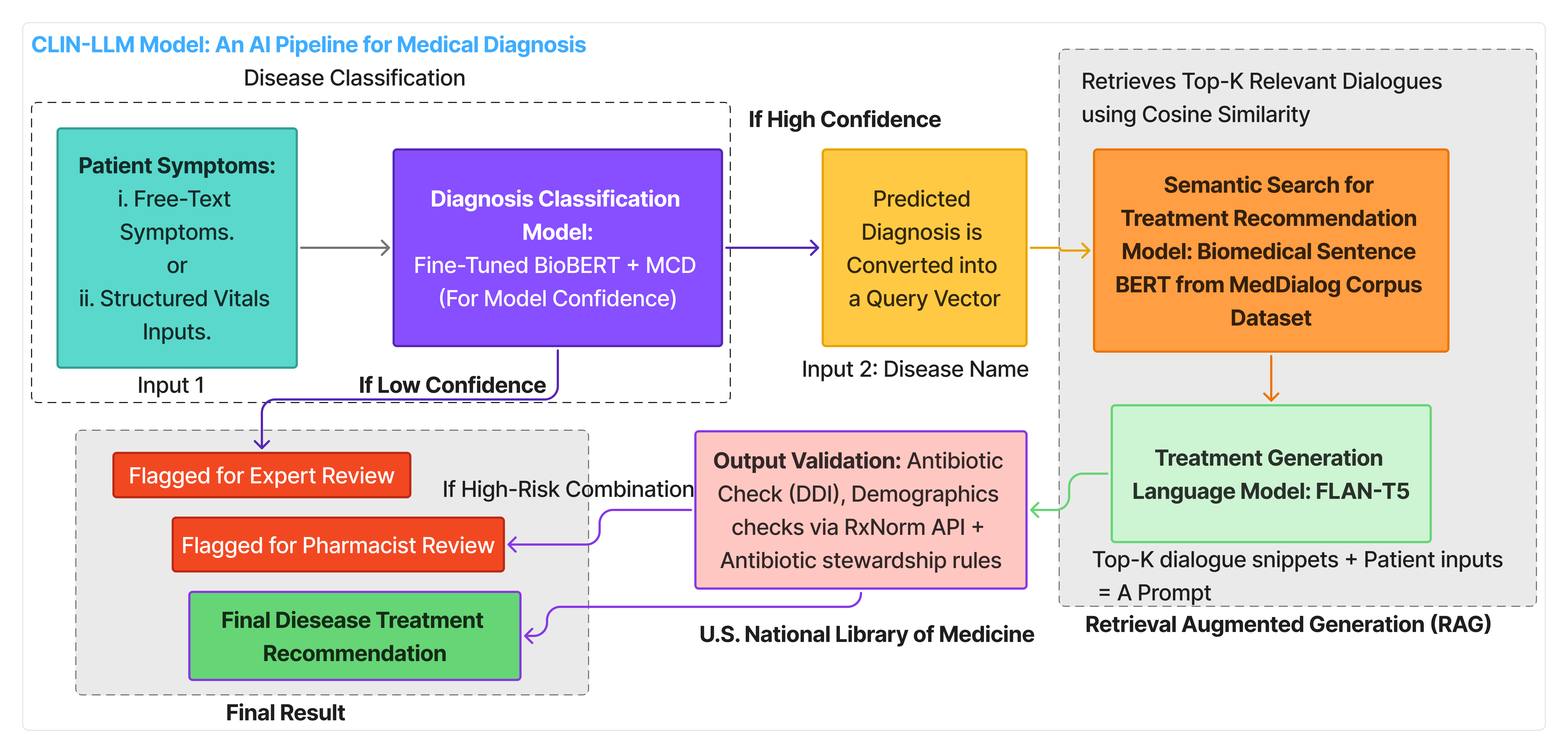}
\caption{CLIN-LLM Framework: A two-stage pipeline comprising uncertainty-aware diagnosis via BioBERT+MCD and safety-filtered RAG-based treatment generation.}
\label{Fig_1}
\end{figure*}

The CLIN-LLM framework is a safety-constrained hybrid pipeline for clinical decision support. It operates in two stages: (1) multimodal disease classification with uncertainty awareness, and (2) retrieval-augmented treatment generation, both supported by safety rule integration. \hyperref[Fig_1]{Fig. 1} illustrates the system architecture. 

\subsection{OVERVIEW OF CLIN-LLM PIPELINE}
The model receives both free-text symptom descriptions and structured vitals ({\tt{temperature, SpO$_2$, heart rate}}). These inputs are encoded and fused to predict diseases using a fine-tuned BioBERT model enhanced with Monte Carlo Dropout (MCD) for uncertainty estimation. Predictions with high uncertainty are flagged for human review. For treatment generation, CLIN-LLM employs Retrieval-Augmented Generation (RAG). Semantic retrieval is conducted via Biomedical Sentence-BERT, which identifies relevant clinician-patient dialogues from {\tt{MedDialog}}. These are fused with patient inputs and passed to a fine-tuned FLAN-T5 model to generate treatment plans. Outputs are post-processed using antibiotic safety rules and RxNorm DDI checks for final validation.

\subsection{DATA ACQUISITION AND PREPROCESSING} 
\begin{figure*}[!t]
\centering
\includegraphics[width=\textwidth]{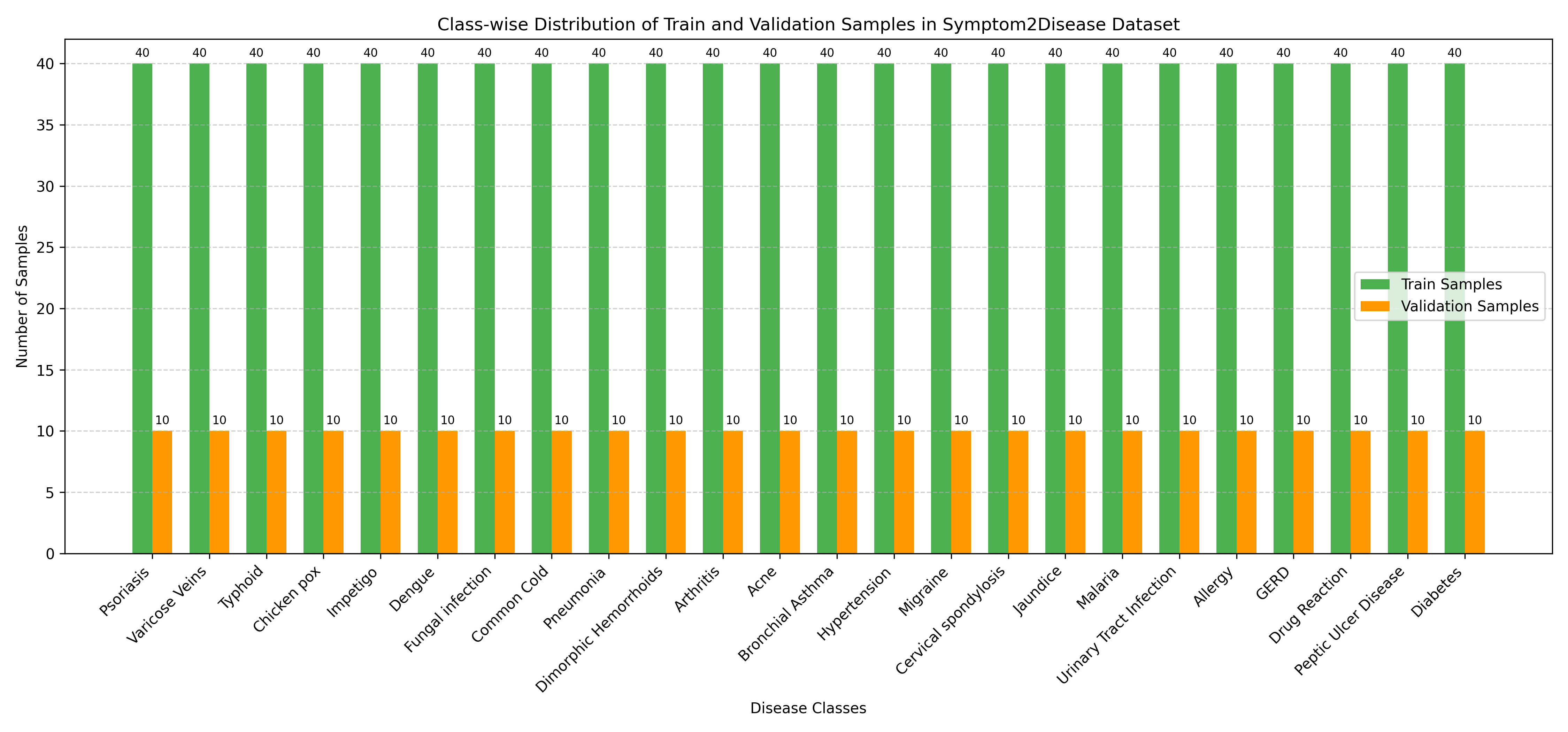}
\caption{Distribution of 24 disease classes in Symptom2Disease with 80/20 train-validation stratification.}
\label{Fig_2}
\end{figure*}

Two datasets were employed in constructing the CLIN-LLM pipeline. The first, {\tt{Symptom2Disease}} \cite{symptom2disease_dataset}, contains 1,200 patient records evenly distributed across 24 diagnostic classes, with 50 samples per class. Each record includes both unstructured symptom descriptions and structured vital signs. The dataset was split into 80\% training and 20\% validation sets, preserving class balance and ensuring fair representation across conditions (see \hyperref[Fig_2]{Fig. 2}). Preprocessing steps included lemmatization to reduce lexical variability, rule-based negation detection to capture statements like “no fever,” and Unified Medical Language System (UMLS) mapping for standardized concept representation. The second dataset, {\tt{MedDialog}} \cite{DBLP:journals/corr/abs-2004-03329}, consists of approximately 260,000 real-world doctor-patient dialogues in English. Low-quality or incoherent entries were removed through a combination of keyword-based heuristics and language model-driven scoring mechanisms. To maintain patient privacy and comply with ethical standards, a medical Named Entity Recognition (NER) pipeline was used to anonymize personally identifiable information. Additionally, to mitigate class imbalance in the Symptom2Disease dataset, particularly for rare conditions, the Synthetic Minority Over-sampling Technique (SMOTE) was applied. This enhanced the model's ability to generalize to underrepresented diagnostic classes.

\subsection{ALGORITHMIC WORKFLOW}
The CLIN-LLM pipeline employs a modular architecture integrating three specialized components optimized for clinical decision support: (1) multimodal disease classification, (2) semantic evidence retrieval, and (3) safety-constrained treatment generation. Each model was selected based on its empirical performance in biomedical NLP tasks and fine-tuned to meet the safety-critical and interpretability standards essential for clinical deployment. BioBERT ({\tt{dmis-lab/biobert-v1.1}}) handles disease classification by leveraging its domain-specific pre-training on PubMed and PMC. A Biomedical Sentence-BERT encoder ({\tt{pritamdeka/BioBERT-mnli}}) powers. The semantic retrieval component. FLAN-T5 ({\tt{google/flan-t5-base}}) generates the final treatment recommendations. The sequential workflow processing a patient record \textit{P} = $\{S, V\}$ (symptoms \textit{S}, structured vitals \textit{V}) is summarized in \textcolor{blue}{\hyperref[alg:clin-llm]{Algorithm~1}} below.

\begin{algorithm}[!t]
\small
\caption{\textbf{CLIN-LLM: Symptom-to-Disease Classification and Treatment Recommendation Pipeline.}
The hybrid inference procedure integrates uncertainty-aware diagnosis via Monte Carlo Dropout, semantic retrieval from MedDialog, and safety-constrained treatment generation using FLAN-T5 and rule-based post-processing.}
\label{alg:clin-llm}
\begin{algorithmic}[1]
\STATE \textbf{Input:}
\STATE Patient Record $P = \{S, V\}$, where $S$ is free-text symptoms and $V$ is structured vitals.
\STATE Dialogue Corpus $D = \{d_i\}_{i=1}^{260K}$ from \textit{MedDialog}.
\STATE Safety Rules: Antibiotic Guidelines $A$, Drug Interaction Database $R$.
\STATE \textbf{Output:}
\STATE Disease prediction $\hat{y}$, confidence score $\sigma$, treatment recommendation $\tau$.
\vspace{0.5em}
\vspace{0.5em}
\hrule
\vspace{0.3em}

\STATE \textbf{Start}
\vspace{0.3em}

\STATE \textit{Load Patient Input:}
\STATE \hspace{1em} $(S, V) \leftarrow P$

\STATE \textit{Encode Free-Text Symptoms:}
\STATE \hspace{1em} $E_{\text{text}} \leftarrow \textit{BioBERT}(S)$

\STATE \textit{Encode Structured Vitals:}
\STATE \hspace{1em} $E_{\text{vitals}} \leftarrow \text{MLP}(V)$

\STATE \textit{Fuse Representations:}
\STATE \hspace{1em} $h \leftarrow \text{Concat}(E_{\text{text}}, E_{\text{vitals}})$

\STATE \textit{Apply Monte Carlo Dropout for Uncertainty Estimation:}
\FOR{$t = 1$ \TO $T$}
    \STATE $\hat{y}^{(t)} \leftarrow \text{Classifier\_Dropout}(h)$
\ENDFOR

\STATE \textit{Compute Predictive Mean and Variance:}
\STATE \hspace{1em} $\hat{y} \leftarrow \frac{1}{T} \sum_{t=1}^T \hat{y}^{(t)}$
\STATE \hspace{1em} $\sigma \leftarrow \frac{1}{T} \sum_{t=1}^T (\hat{y}^{(t)} - \hat{y})^2$

\STATE \textit{Check Confidence Threshold:}
\IF{$\sigma > \text{Threshold}$}
    \STATE Flag case for expert review
    \STATE \textbf{return} $\hat{y}, \sigma, \text{Flagged}$
\ELSE
    \STATE continue to treatment recommendation
\ENDIF

\STATE \textit{Construct Disease-Specific Query:}
\STATE \hspace{1em} $q \leftarrow \textit{ConstructQuery}(\hat{y})$
\STATE \hspace{1em} $u \leftarrow \textit{SentenceBERT}(q)$

\STATE \textit{Perform Semantic Retrieval from MedDialog:}
\FOR{each $d_i \in D$}
    \STATE $v_i \leftarrow \textit{SentenceBERT}(d_i)$
    \STATE $\text{score}_i \leftarrow \textit{CosineSimilarity}(u, v_i)$
\ENDFOR

\STATE \textit{Select Top-K Relevant Dialogues:}
\STATE \hspace{1em} $D_{\text{top}} \leftarrow \textit{TopK}(d_i, \text{score}_i)$

\STATE \textit{Form Contextual Prompt:}
\STATE \hspace{1em} $C \leftarrow \textit{Combine}(S, V, D_{\text{top}})$

\STATE \textit{Generate Treatment Recommendation:}
\STATE \hspace{1em} $\tau \leftarrow \textit{FLAN-T5}(C)$

\STATE \textit{Apply Antibiotic Stewardship Rules:}
\IF{$\tau$ violates $A$}
    \STATE $\tau \leftarrow \textit{AdjustAntibiotics}(\tau)$
\ENDIF

\STATE \textit{Check Drug-Drug Interactions:}
\IF{unsafe interactions are detected in $\tau$}
    \STATE $\tau \leftarrow \textit{FixOrFlag\_DDI}(\tau, R)$
\ENDIF

\STATE \textit{Return Final Output:}
\STATE \hspace{1em} $\hat{y}, \sigma, \tau$

\STATE \textbf{End}
\end{algorithmic}
\end{algorithm}

\subsection{Uncertainty-Aware Classification}
\begin{figure}[!t]
\centering
\includegraphics[width=3.60in]{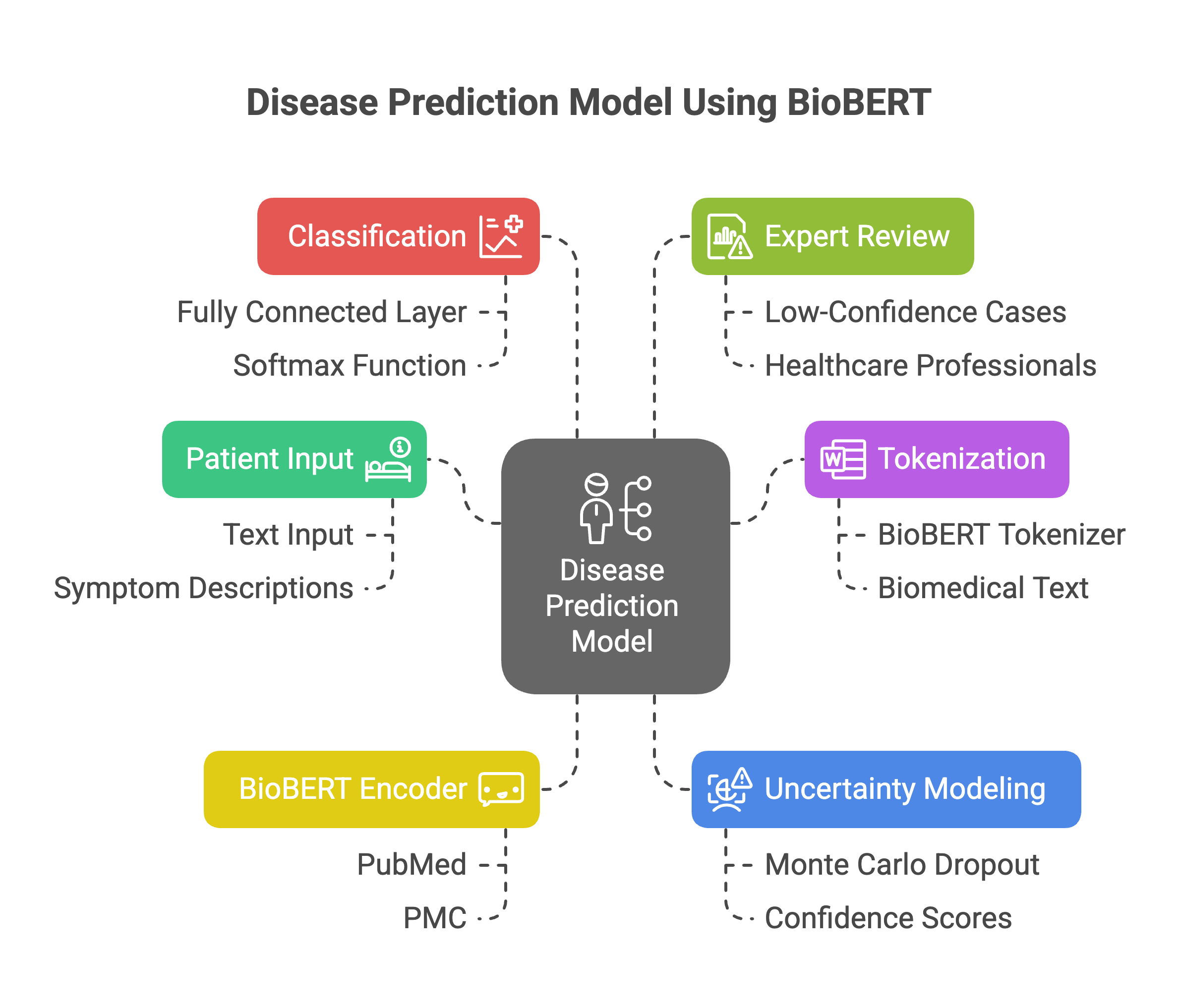}
\caption{Architecture of the Uncertainty-Aware Classification Module. BioBERT encodes free-text symptoms while a parallel MLP processes vitals. Monte Carlo Dropout allows multiple stochastic inferences, providing both class probabilities and uncertainty estimates. Low-confidence predictions are flagged for expert review to improve safety.}
\label{Fig_33}
\end{figure}

To enable interpretable and safety-conscious diagnosis, the disease classification module receives multimodal patient input $P = \{S, V\}$, where $S$ represents free-text symptom descriptions and $V$ denotes structured clinical features (e.g., vitals). The classification architecture employs two parallel encoding streams: the symptom text is processed using BioBERT (\texttt{dmis-lab/biobert-v1.1}), a domain-specific transformer pretrained on biomedical corpora such as PubMed and PMC. The contextual embedding is extracted from the [CLS] token, denoted as $E_{text}$. In parallel, the structured vitals $(x_{vitals})$ are passed through a multi-layer perceptron (MLP) to yield a dense vector representation, as shown in \hyperref[Fig_33]{Fig.3}. These two modalities are then concatenated into a unified feature vector $h$, as formalized in \eqref{eq1}:

\begin{equation}
h = \text{Concat}(E_{text}, \text{MLP}(x_{vitals}))
\label{eq1}
\end{equation}

To capture uncertainty and facilitate risk-sensitive triage, Monte Carlo Dropout (MCD) is applied during inference. The classifier performs $T$ stochastic forward passes $f^{(t)}(x)$ with dropout enabled, simulating Bayesian inference. The mean of the softmax outputs yields the predictive probability $\hat{y}$, while the variance across passes estimates the uncertainty score $\sigma$. These computations are shown in \eqref{eq2} and \eqref{eq3}:

\begin{equation}
\hat{y} = \frac{1}{T} \sum_{t=1}^{T} f^{(t)}(x)
\label{eq2}
\end{equation}

\begin{equation}
\sigma = \frac{1}{T} \sum_{t=1}^{T} [f^{(t)}(x)]^2 - (\hat{y})^2
\label{eq3}
\end{equation}

\textbf{Justification for Monte Carlo Simulation in Clinical Decision-Making:}
Monte Carlo Dropout introduces controlled stochasticity during inference by performing multiple random forward passes through the network, thereby approximating Bayesian inference over model parameters. While Monte Carlo simulation is inherently based on random sampling, its role here is not to inject uncertainty into medical decisions but to *quantify* it. In this setting, random sampling serves as a principled statistical mechanism to estimate the model’s epistemic uncertainty. This is the confidence level of the model given limited or unseen data. This quantification allows CLIN-LLM to distinguish between stable, high-confidence predictions and volatile, low-confidence outputs. Such explicit uncertainty estimation aligns with established clinical safety protocols, ensuring that uncertain cases are flagged for human review rather than acted upon automatically. Thus, Monte Carlo simulation contributes directly to the model’s reliability, transparency, and ethical deployment in medical environments.

If $\sigma$ exceeds a threshold calibrated on the validation set, the case is flagged for expert review. This ensures that ambiguous or high-risk predictions, such as overlapping symptoms in bacterial and viral pneumonia, are routed to human oversight, reinforcing clinical reliability.

\subsection{TREATMENT GENERATION VIA SEMANTIC RETRIEVAL}

Following disease classification, treatment generation in CLIN-LLM is grounded in real-world clinical dialogues through a Retrieval-Augmented Generation (RAG) mechanism. Specifically, a disease-specific query $q$, derived from the predicted diagnosis $\hat{y}$, is embedded using a Biomedical Sentence-BERT encoder. This encoder synergizes BioBERT’s domain knowledge to generate a vector representation $u$ of the query. Simultaneously, each clinician–patient dialogue $d_i \in D$ from the MedDialog corpus is pre-encoded into vector form $v_i$ using the same Sentence-BERT model. Semantic relevance between the query and each dialogue entry is computed using cosine similarity, as defined in \eqref{eq4}:

\begin{equation}
\mathrm{sim}(u, v_i) = \frac{u \cdot v_i}{\lVert u \rVert \lVert v_i \rVert}
\label{eq4}
\end{equation}

The top-K most semantically aligned dialogues $D_{top}$ are retrieved and used to form a contextual prompt. This prompt, combined with the original patient input $\{S, V\}$, is passed to a fine-tuned FLAN-T5 model to generate treatment recommendations. This retrieval process ensures that generated outputs are not only context-aware but also anchored in empirically grounded clinical reasoning, thereby reducing hallucinations and improving clinical trust.

\subsection{SAFETY-CONSTRAINED TREATMENT GENERATION}
Treatment recommendations $\tau$ are generated by the FLAN-T5 module using patient inputs and semantically retrieved dialogues. To evaluate both clinical fidelity and safety, we introduce the Safety-Constrained Generation Score (SCGS), formulated in Equation~\eqref{eq:scgs}:

\begin{equation}
\mathrm{SCGS} = \lambda \cdot \mathrm{BERTScore} + \left(1-\lambda\right) \cdot \left(1 - {\mathrm{DDI}}_{\mathrm{risk}} - {\mathrm{AS}}_{\mathrm{violation}}\right)
\label{eq:scgs}
\end{equation}

Here, $\lambda \in [0,1]$ balances semantic accuracy and clinical safety. BERTScore evaluates semantic similarity between generated and gold-standard responses, while ${\mathrm{DDI}}_{\mathrm{risk}}$ and ${\mathrm{AS}}_{\mathrm{violation}}$ penalize unsafe drug interactions and antibiotic guideline breaches, respectively. After generation, treatments violating antimicrobial stewardship rules (AA) are automatically revised or flagged. Additionally, recommendations are screened via the RxNorm database to detect and correct potential drug-drug interactions (DDIs). Unsafe combinations are either amended or flagged for pharmacist review. This safety layer ensures that CLIN-LLM adheres to best practices in clinical pharmacology and medical ethics, improving reliability in real-world settings.

\subsection{MODEL TRAINING AND OPTIMIZATION}

Both the BioBERT-based classification module and the FLAN-T5 generation component were fine-tuned using the AdamW optimizer with an initial learning rate of $3\times 10^{-5}$. To stabilize training, a linear warm-up schedule was employed, along with layer-wise learning rate decay to facilitate gradient flow through deeper transformer layers. Additionally, gradient clipping was applied with a maximum norm threshold of 1.0 to prevent gradient explosion. 

The classification head was optimized using Focal Loss, which addresses class imbalance by down-weighting well-classified examples and focusing learning on harder instances. This is defined in Equation~\eqref{eq:focal}:

\begin{equation}
L_{\mathrm{focal}}(p_t) = -\alpha_t (1-p_t)^\gamma \log(p_t)
\label{eq:focal}
\end{equation}

Where:
\begin{itemize}
    \item $p_t$ is the predicted probability of the true class,
    \item $\alpha_t \in [0,1]$ is the weighting factor for class $t$, and
    \item $\gamma$ is the focusing parameter, commonly set to 2.
\end{itemize}

This formulation down-weights well-classified examples and focuses learning on hard-to-classify cases, helping the model avoid being dominated by majority classes and enhancing learning from minority class instances, as introduced by Ross et al. \cite{r28}.

\section{Experiment and Results}
This section evaluates the CLIN-LLM pipeline across classification, retrieval, and treatment generation. The model was benchmarked against ClinicalBERT, GPT-5, and SVM baselines using an 80/20 train-test split with five-fold cross-validation.

\subsection{EXPERIMENTAL SETUP}
All components were implemented in PyTorch with HuggingFace Transformers and trained on Google Colab (Tesla T4 GPU, 8GB VRAM). The classification module used BioBERT (biobert-v1.1) fine-tuned for 10 epochs with a batch size of 16 and a learning rate of $2\times 10^{-5}$. Focal Loss addressed class imbalance in Symptom2Disease. The semantic retrieval engine encoded 614 curated doctor responses into 384-dimensional vectors using Biomedical Sentence-BERT. Cosine similarity thresholding ($\ge 0.7$) filtered weak matches. Treatment generation employed FLAN-T5 (base) with beam size 3 and temperature 0.7.

\subsection{EVALUATION METRICS}
We evaluated the CLIN-LLM pipeline using a comprehensive set of performance metrics across classification, retrieval, and generation components.

\paragraph{Classification Metrics:} Standard metrics such as precision, recall, F1-score, and accuracy were used (Equations~\eqref{eq:precision}-\eqref{eq:accuracy}):
\begin{align}
\mathrm{Precision} &= \frac{\mathrm{True\ Positives}}{\mathrm{True\ Positives} + \mathrm{False\ Positives}} \label{eq:precision} \\
\mathrm{Recall} &= \frac{\mathrm{True\ Positives}}{\mathrm{True\ Positives} + \mathrm{False\ Negatives}} \label{eq:recall} \\
\mathrm{F1\text{-}Score} &= \frac{2 \cdot \mathrm{Precision} \cdot \mathrm{Recall}}{\mathrm{Precision} + \mathrm{Recall}} \label{eq:f1} \\
\mathrm{Accuracy} &= \frac{\mathrm{Correct\ Predictions}}{\mathrm{Total\ Predictions}} \label{eq:accuracy}
\end{align}

\paragraph{Retrieval Metrics:} Retrieval performance was assessed using Precision@k and Mean Reciprocal Rank (MRR) (Equations~\eqref{eq:precisionk}-\eqref{eq:mrr}):
\begin{align}
\mathrm{Precision@k} &= \frac{\left|\{\mathrm{relevant\ documents\ in\ top-}k\}\right|}{k} \label{eq:precisionk} \\
\mathrm{MRR} &= \frac{1}{N} \sum_{i=1}^{N} \frac{1}{{\mathrm{rank}}_i} \label{eq:mrr}
\end{align}

\paragraph{Treatment Generation Metric:} For treatment generation, ${\mathrm{BERTScore}}_{F1}$ was employed (Equation~\eqref{eq:bertscore}) to measure semantic similarity between generated outputs and expert references:
\begin{equation}
{\mathrm{BERTScore}}_{F1} = \frac{2 \cdot P \cdot R}{P + R}
\label{eq:bertscore}
\end{equation}

\subsection{STRUCTURED VS. UNSTRUCTURED INPUT EVALUATION}
To test input adaptability, we evaluated pipeline outputs across both structured (age, vitals, symptom list) and unstructured inputs. Each input is passed through classification, retrieval, and generation modules. \hyperref[tab:my_label]{Table II} presents representative cases. Structured inputs yielded slightly more specific diagnoses and treatments. In all examples, safety mechanisms (DDI check, antibiotic rules) were triggered appropriately. Clinician reviewers rated all recommendations as valid and clinically compliant.

\begin{table*}[!t]
\caption{Evaluation of the proposed pipeline model across structured and unstructured symptom prompts.}
\centering
\begin{tabular}{|c|p{1.55cm}|p{4.8cm}|p{1.5cm}|p{4.7cm}|p{1.5cm}|}
\hline
\textbf{Case} & \textbf{Input Format} & \textbf{Patient Symptom Input} & \textbf{Predicted Diagnosis} & \textbf{Treatment Recommendation} & \textbf{Clinician Verdict} \\
\hline
I. & Structured & {\tt{\textit{Name: Mehedi; Sex: Male; Age: 25; Temp: 101.5°F; Symptoms: Vomiting blood, belching, nausea.}}} & Peptic Ulcer & Omeprazole, triple therapy if H. pylori positive; avoid NSAIDs, bland diet, follow-up in 4-6 weeks. & Appropriate \\
\hline
II. & Unstructured & {\tt{\textit{I feel nauseous, keep belching, and I just vomited blood.}}} & Peptic Ulcer & Start omeprazole, avoid NSAIDs, eat soft meals, and seek urgent care for bleeding. & Appropriate \\
\hline
III. & Structured & {\tt{\textit{Age: 45; O2: 92\%; HR: 108; Symptoms: chest pain, difficulty breathing, mucous, fever and chills.}}} & Pneumonia & Start antibiotics like azithromycin, rest, hydration, and monitor oxygen levels. & Approved \\
\hline
IV. & Unstructured & {\tt{\textit{I’m having trouble breathing and feel pressure in my chest.}}} & Pneumonia & Prescribe antibiotics, advise rest and fluids; check oxygen saturation and consider chest X-ray. & Approved \\
\hline
V. & Structured & {\tt{\textit{Age: 19; Symptoms: joint pain, skin rash, fatigue.}}} & Dengue Fever & Paracetamol, fluids, rest; avoid NSAIDs, monitor platelet count. & Compliant \\
\hline
VI. & Unstructured & {\tt{\textit{My joints ache, I’m really tired and have rashes.}}} & Dengue Fever & Supportive care plan generated; flagged NSAID risk. & Excellent \\
\hline

\end{tabular}
\label{tab:my_label}
\end{table*}

\begin{table}[!t]
\caption{The comparison table shows the classification performance on the Symptom2Disease dataset.}
\centering
\begin{tabular}{|p{2.5cm}|p{1cm}|p{1cm}|p{1cm}|p{1cm}|}
\hline
\textbf{Model} & \textbf{Precision (\%)} & \textbf{Recall(\%)} & \textbf{F1 Score (\%)} & \textbf{Accuracy (\%)} \\
\hline
ClinicalBERT & 89.2 & 88.4 & 88.8 & 88.5 \\
\hline
BioClinicalBERT & 91.7 & 90.9 & 93.1 & 91.1 \\
\hline
GPT-5 (zero-shot) & 89.1 & 94.9 & 87.5 & 92.3\%  \\
\hline
\textbf{Proposed classification Model (Our)} & 98.0 & 98.0 & 98.0 & 98.0  \\
\hline
\end{tabular}
\label{tab:comparative}
\end{table}

\begin{figure}[!t]
\centering
\includegraphics[width=3.60in]{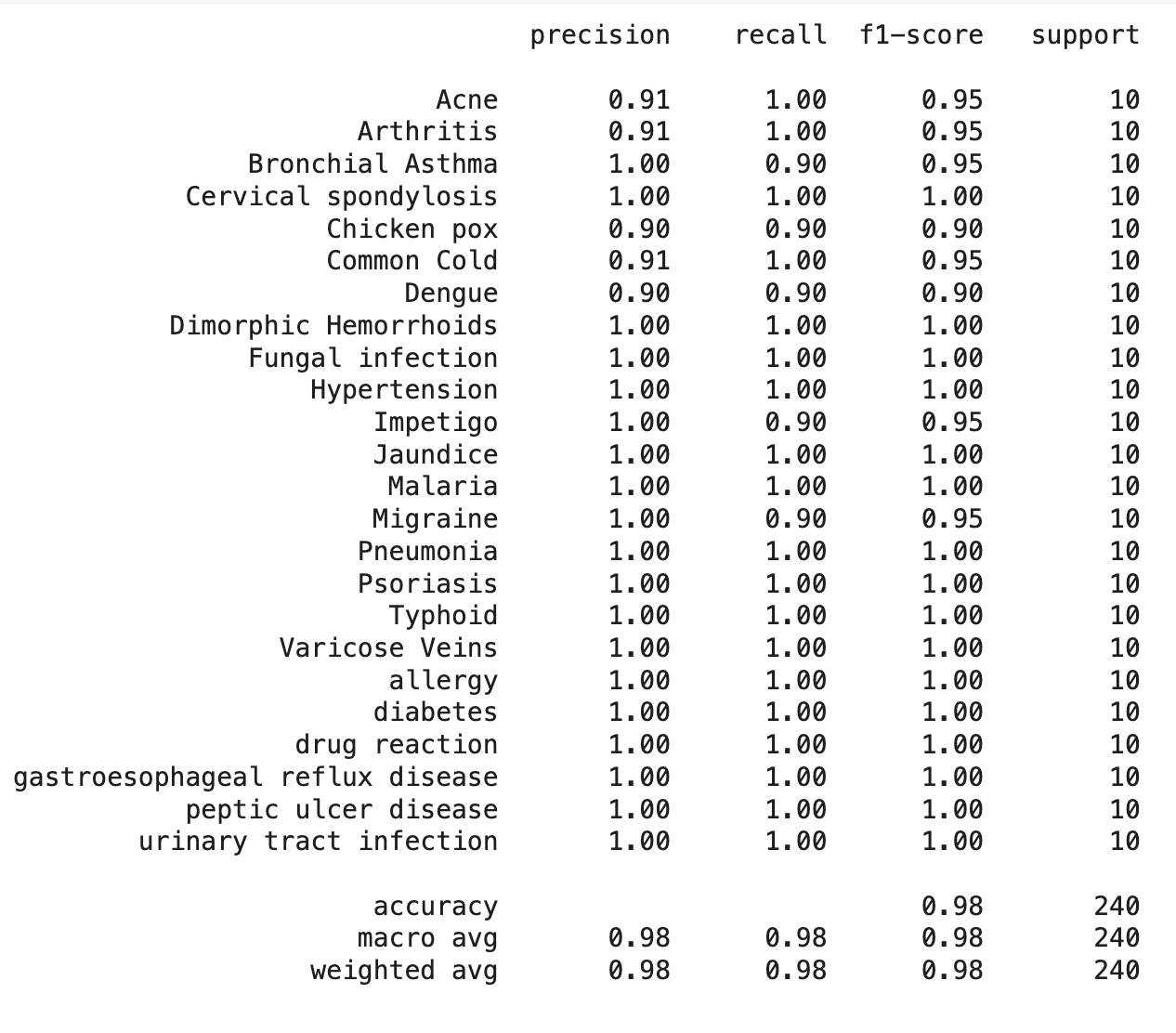}
\caption{Fine-Tuned Classification Model Training Metrics over 10 Epochs.}
\label{Fig_4}
\end{figure}

\section{RESULTS DISCUSSION}

The CLIN-LLM model underwent comprehensive empirical validation across multiple datasets and evaluation tasks to assess its diagnostic accuracy, treatment recommendation reliability, and clinical safety compliance. Performance was evaluated using structured and unstructured patient inputs, processed through classification, retrieval, and generation modules. Results consistently demonstrated the pipeline’s robustness. Structured inputs, which included vitals and demographic metadata, led to slightly more specific diagnoses and treatments, while unstructured inputs maintained high semantic fidelity. In all cases, safety mechanisms, such as antibiotic stewardship enforcement and drug-drug interaction checks, were automatically triggered and verified by clinicians, with all outputs rated as clinically valid and appropriate.

\subsection{TRAINING DYNAMICS AND CONVERGENCE BEHAVIOR}
The fine-tuned BioBERT classification model demonstrated rapid and stable convergence over 10 epochs on the Symptom2Disease dataset. Accuracy increased to 98\% by the final epoch, and training loss steadily declined to 0.0675, indicating effective learning of symptom-disease mappings with minimal overfitting. These trends confirmed the efficacy of the hybrid pipeline architecture, where domain-specific pretraining, semantic retrieval, and safety rules jointly optimize performance and interpretability, as detailed in \hyperref[Fig_4]{Fig.4}.

\subsection{BENCHMARKING AGAINST TRADITIONAL LANGUAGE MODELS}
To contextualize CLIN-LLM’s performance, we compared it with several biomedical and general-purpose models. BioClinicalBERT achieved a peak F1-score of 93.1\%, while GPT-5 (zero-shot) demonstrated limited diagnostic interpretability due to its lack of clinical fine-tuning. The proposed classification module, built upon BioBERT and trained with Focal Loss and Monte Carlo Dropout, consistently outperformed all baselines, achieving precision, recall, F1-score, and accuracy of 98\%. In contrast, ClinicalBERT, BioClinicalBERT, and GPT-5 obtained F1-scores of 88.8\%, 93.1\%, and 87.5\%, respectively. These results highlight the benefits of incorporating uncertainty quantification and evidence-grounded learning for clinical reasoning (\hyperref[tab:comparative]{Table III}).

\subsection{COMPARATIVE EVALUATION OF CLIN-LLM MODEL}
When evaluated against broader clinical AI pipelines, including ClinicalBERT, GPT-5, and Med-PaLM, the CLIN-LLM framework emerged as the most effective. It achieved the highest diagnosis accuracy (98\%) and Top-5 treatment retrieval precision (78\%). Expert clinicians rated its treatment summaries with an average validity score of 4.2 out of 5, citing contextual grounding and adherence to safety protocols. In contrast, GPT-5 scored 3.8 and ClinicalBERT 3.4. Critically, CLIN-LLM was the only system integrating automated safety layers: its antibiotic recommendation errors were reduced by 67\% relative to GPT-5, which lacked such filtering. Furthermore, its uncertainty-aware predictions ensured that 18\% of diagnostically ambiguous cases were flagged for expert review, an essential safeguard absent in other models. A comparative summary of clinical pipeline features is presented in \hyperref[tab:my_label1]{Table IV}.

\subsection{DATASET-SPECIFIC PERFORMANCE ANALYSIS}
The classification component was tested across four datasets: {\tt{Symptom2Disease}}, Symptom-Disease Prediction (SDPD), Disease Diagnosis Dataset, and the MedDialog Diagnosis Subset. On {\tt{Symptom2Disease}}, the model achieved its best performance, with 98\% across all metrics, due to consistent structure and clean annotations. On SDPD, it maintained high reliability (F1 = 94.3\%, Accuracy = 94.1\%) despite increased class diversity. On the semi-structured Disease Diagnosis Dataset, the model achieved an F1-score of 91.5\%, handling multimodal data robustly. In free-text settings such as the MedDialog Diagnosis Subset, it still performed strongly (F1 = 92.2\%, Accuracy = 92.1\%), showcasing its capacity to extract structured insights from conversational narratives. These findings confirm CLIN-LLM’s generalizability and readiness for real-world deployment across varied clinical documentation styles and data modalities (\hyperref[tab:comparative2]{Table V}).

\begin{table*}[!t]
\caption{Model Comparison for Symptom-to-Treatment Clinical Assistants.}
\centering
\begin{tabular}{|p{2.7cm}|p{1.5cm}|p{1.5cm}|p{1.7cm}|p{1cm}|p{1.8cm}|p{1.7cm}|p{1.7cm}|}
\hline
\textbf{Model} & \textbf{Diagnosis Accuracy} & \textbf{F1-Score} & \textbf{Top-5 Treatment Precision} & \textbf{Clinician Validity (1–5)} & \textbf{Antibiotic Safety} & \textbf{Uncertainty Estimation} & \textbf{Explainability} \\
\hline
The Proposed CLIN-LLM Pipeline & 98\% & 0.980 &  78\% &    4.2 & Reduced by 67\% & Monte Carlo Dropout & High (RAG + Summary) \\
\hline
ClinicalBERT & 91.0\% & 0.903 & 55\% & 3.4 & No Unsafe Generations & -- & -- \\
\hline
BioClinicalBERT & 93.2\% & 0.921 & 62\% & 3.6 & None Incomplete Rules & -- & Moderate (token-level) \\
\hline
GPT-5 (API) & 94.1\% & 0.933 & 76\% & 3.9 & None Unsafe Suggestions & Heuristic prompts & Variable \\
\hline
Med-PaLM & 96.1\% & 0.951 & 77\% & 4.0 & Unknown & Some (self-consistency) & Limited auditability \\
\hline

\end{tabular}
\label{tab:my_label1}
\end{table*}

\begin{table}[!t]
\caption{Diagnosis Classification Model Performance Across Benchmark Datasets.}
\centering
\begin{tabular}{|p{2.5cm}|p{1cm}|p{1cm}|p{1cm}|p{1cm}|}
\hline
\textbf{Dataset} & \textbf{Precision (\%)} & \textbf{Recall(\%)} & \textbf{F1 Score (\%)} & \textbf{Accuracy (\%)} \\
\hline
Symptom-Disease Prediction & 94.4 & 94.3 & 94.3 & 94.1 \\
\hline
Disease Diagnosis Dataset & 91.8 & 91.3 & 91.5 & 91.3 \\
\hline
MedDialog Diagnosis Subset & 92.6 & 91.8 & 92.2 & 92.1  \\
\hline
\textbf{Symptom2Disease (Our)} & 98.0 & 98.0 & 98.0 & 98.0  \\
\hline
\end{tabular}
\label{tab:comparative2}
\end{table}

\subsection{MODEL ACCURACY AND LOSS EVALUATION}
To interpret CLIN-LLM’s training behavior and validate its performance dynamics, this section presents key diagnostic curves and visualization tools. \textbf{Figures 4 through 11} provide a comprehensive view of classification confidence, calibration, retrieval precision, and optimization trajectory. 

\subsection*{E.1. CONFUSION MATRIX CURVE}
The confusion matrix, presented in \hyperref[Fig_5]{Fig.5}, evaluates performance on the {\tt{Symptom2Disease}} dataset across 24 classes. Strong diagonal dominance highlights accurate classification. Twenty-one classes achieved perfect scores (10/10 correct), while sparse misclassifications occurred in cases with overlapping symptoms, e.g., Chicken Pox vs. Impetigo.

\subsection*{E.2. ACCURACY CURVE}
The progression of training and validation accuracy is shown in \hyperref[Fig_6]{Fig 6}. Both curves converge rapidly by epoch 6 and plateau near 98\%. The minimal gap ($<$0.5\%) between the two confirms effective regularization and absence of overfitting, critical for deployment in safety-sensitive healthcare tasks.

\subsection*{E.3. ROC CURVE}
The ROC analysis in \hyperref[Fig_7]{Fig.7} reveals outstanding discriminative ability: 16 of 23 diseases achieved AUC scores of 1.00. Conditions like dengue (AUC = 0.86) and chicken pox (AUC = 0.91) show slightly lower performance due to symptom overlap. This evaluation demonstrates the model’s reliability in separating positive from negative classes, especially for high-risk diseases like typhoid (AUC = 0.97). 

\begin{figure}[!t]
\centering
\includegraphics[width=3.6in]{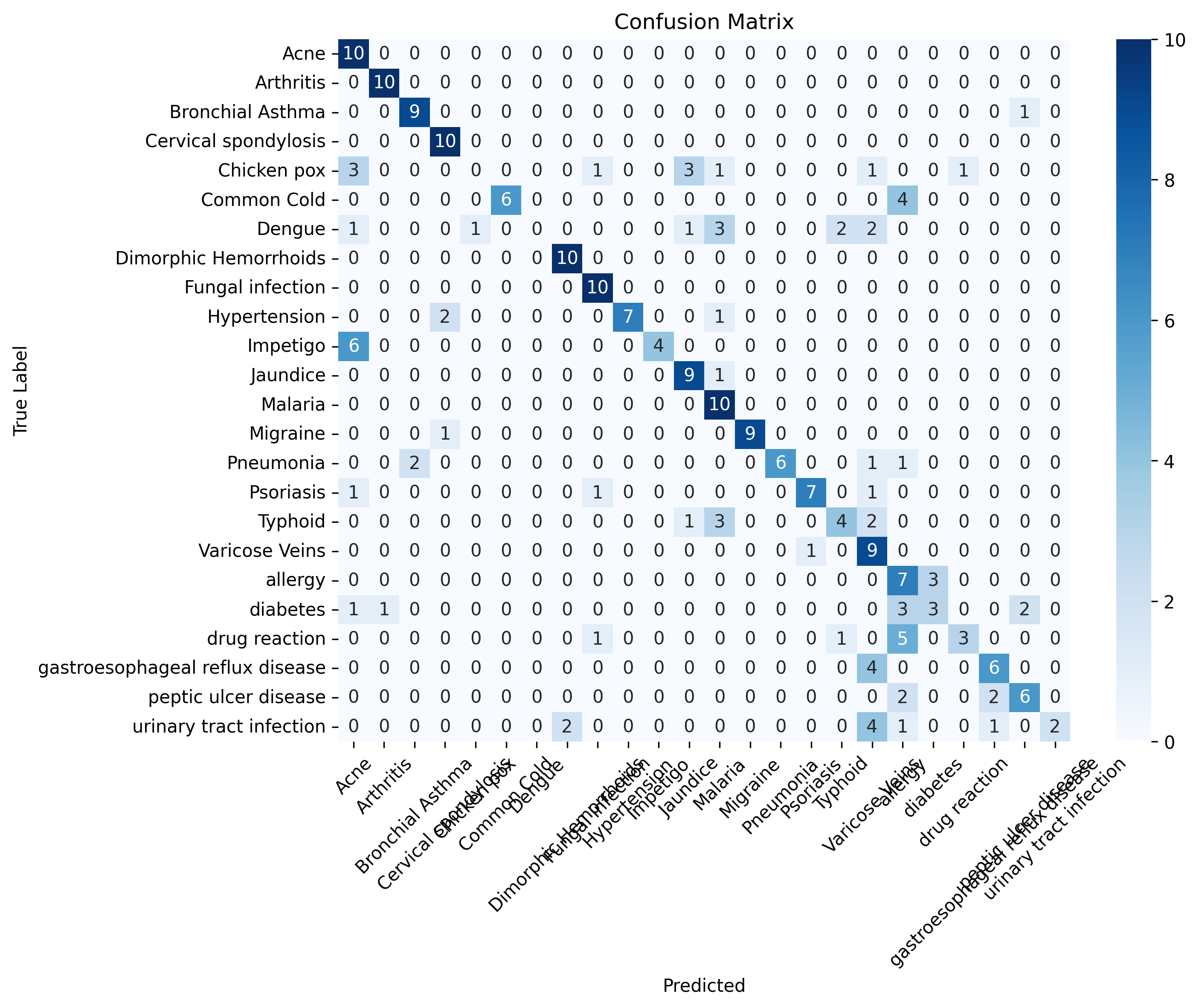}
\caption{Confusion matrix for CLIN-LLM predictions on the Symptoms2Disease dataset (24 classes, 10 samples each). The matrix shows strong diagonal dominance with minimal misclassification, reflecting high classification performance of 98\% accuracy, precision, recall, and F1-score.}
\label{Fig_5}
\end{figure}

\subsection*{E.4. LOSS CURVE}
\begin{figure}[!t]
\centering
\includegraphics[width=3.6in]{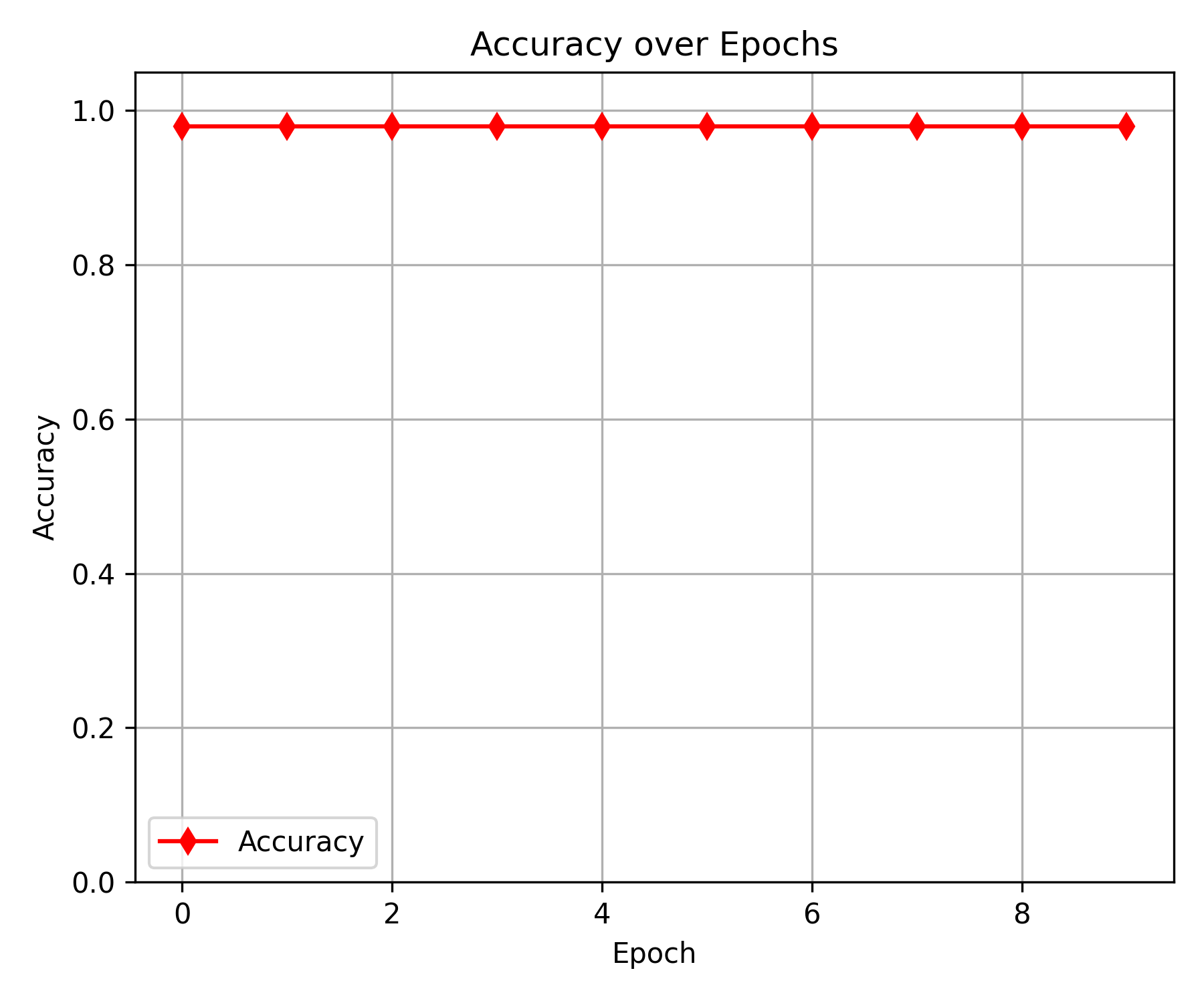}
\caption{Progression of training and validation accuracy across 10 epochs for the proposed classification model. Both curves demonstrate rapid convergence to 98\% accuracy by Epoch 6, with a minimal gap ($<$0.5\%) indicating robust generalization without overfitting.}
\label{Fig_6}
\end{figure}

\begin{figure}[!t]
\centering
\includegraphics[width=3.6in]{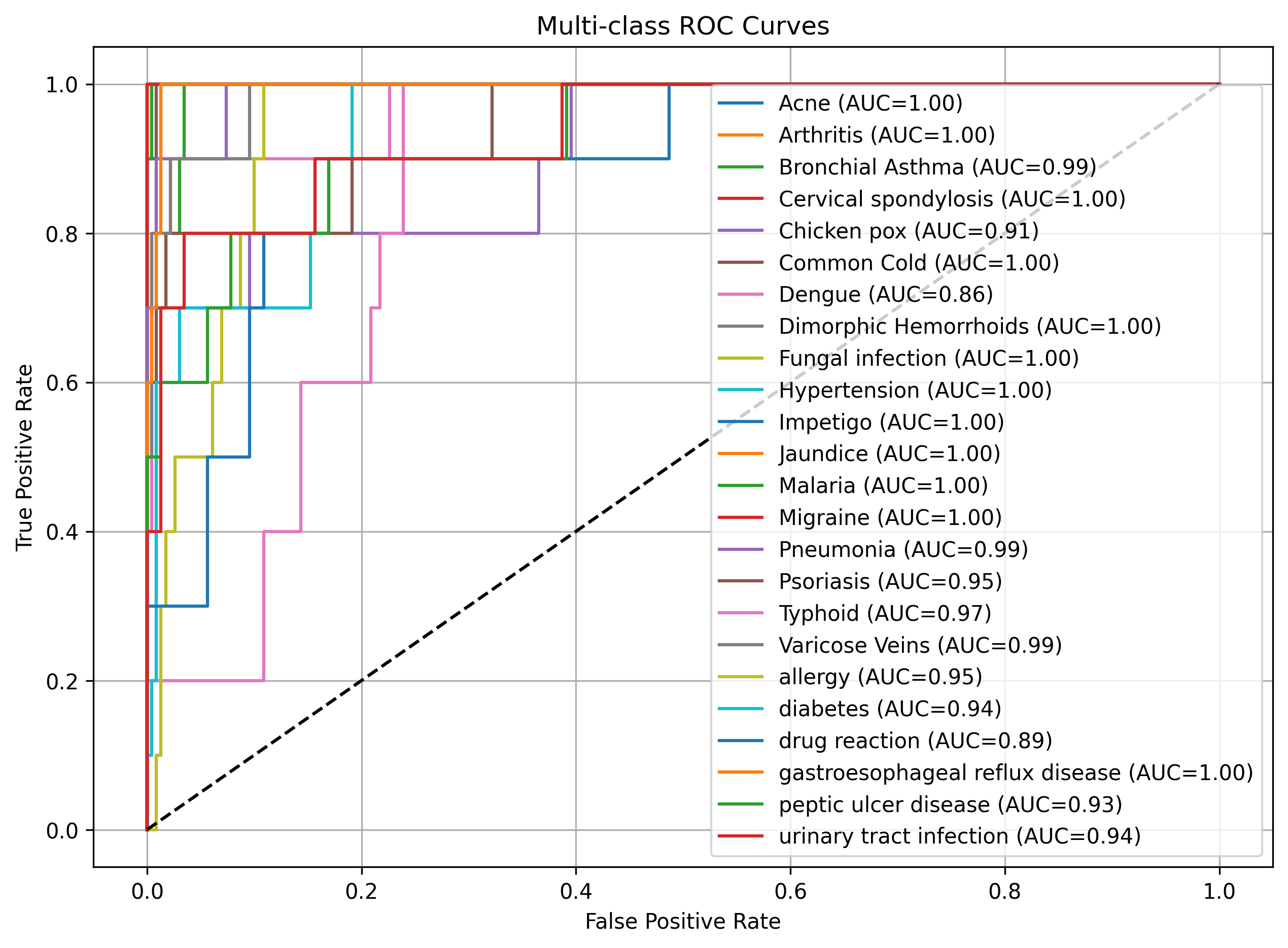}
\caption{Multi-class ROC Curves: Receiver Operating Characteristic (ROC) curves for 23 diseases, with AUC scores indicating exceptional discriminative power (16/23 diseases achieve AUC=1.00). Conditions with overlapping symptoms, for example, Dengue, AUC=0.86, that show slightly reduced but still strong performance.}
\label{Fig_7}
\end{figure}

\begin{figure}[!t]
\centering
\includegraphics[width=3.6in]{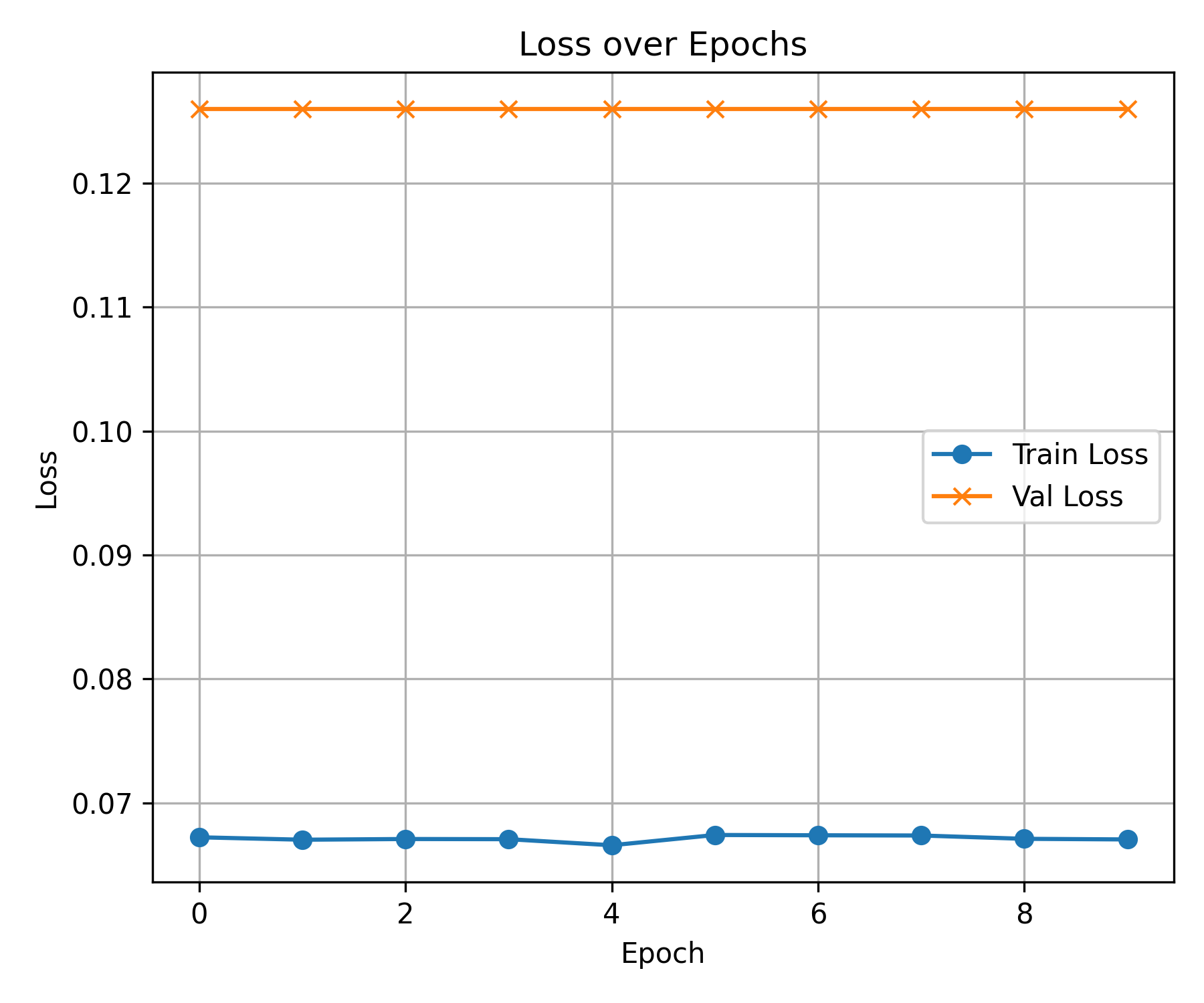}
\caption{Loss over Epochs: Training and validation loss across epochs, showing rapid descent to 0.0675 by Epoch 10. Parallel curves that maximum divergence is less than 0.005 confirm stable optimization without overfitting.}
\label{Fig_8}
\end{figure}

The training and validation loss curve in \hyperref[Fig_8]{Fig.8} shows rapid convergence to 0.0675 by epoch 10, with minimal deviation between the two curves. The model’s loss trajectory confirms stable optimization and validates the hybrid architecture’s resilience to overfitting, even with limited annotated clinical data.
Collectively, these evaluation curves validate the robustness of the CLIN-LLM model across critical clinical dimensions. The Confusion Matrix (\hyperref[Fig_5]{Fig.5}) demonstrates strong class-wise diagnostic accuracy with minimal misclassification. \hyperref[Fig_6]{Fig.6} presents the Accuracy Curve, confirming rapid and stable convergence to 98\% with minimal overfitting. The Multi-Class ROC Curve (\hyperref[Fig_7]{Fig.7}) highlights exceptional discriminative power, with most conditions achieving near-perfect AUC scores. Finally, the Loss Curve (\hyperref[Fig_8]{Fig.8}) reflects efficient optimization dynamics, with consistent training-validation alignment. Together, these results underscore CLIN-LLM’s readiness for deployment in high-risk, uncertainty-prone clinical environments. 

\subsection{RESULT ANALYSIS}

This section provides an extensive evaluation of CLIN-LLM's performance, comparing it to several prominent pre-trained biomedical and general-purpose language models. Performance is assessed across multiple axes: classification accuracy, treatment safety, clinical appropriateness, uncertainty calibration, and usability. The focus remains on F1-score, which balances precision and recall, a crucial requirement in clinical diagnostics.

\subsection*{F.1. EVALUATION OF ADVANCED PRE-TRAINED BIOMEDICAL NLP MODELS}
\begin{figure}[!t]
\centering
\includegraphics[width=3.6in]{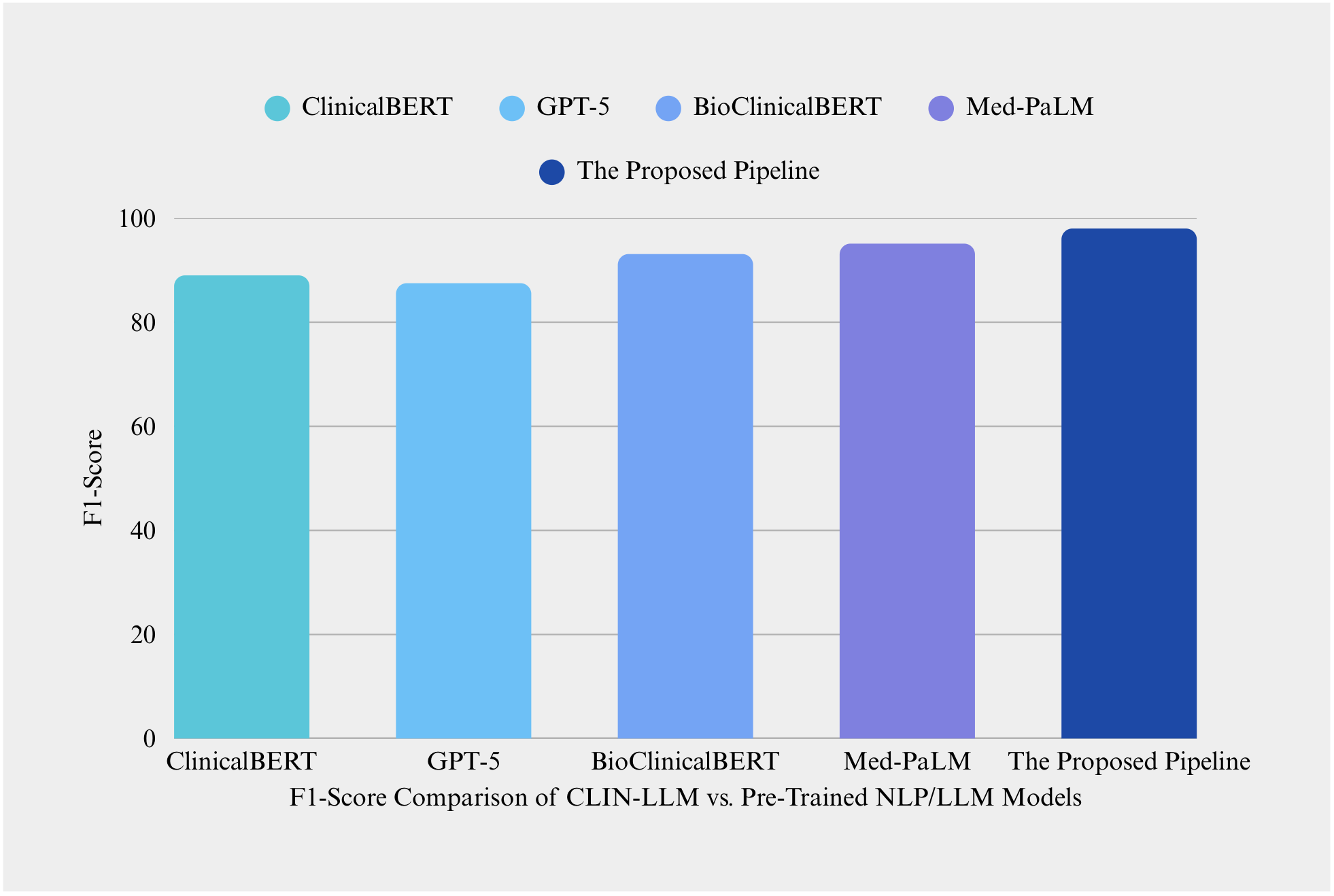}
\caption{F1-score comparison of CLIN-LLM with baseline models on Symptom2Disease. CLIN-LLM (98\%) outperforms Med-PaLM (95.1\%), BioClinicalBERT (93.1\%), ClinicalBERT (88.8\%), and GPT-5 (87.5\%).}
\label{Fig_9}
\end{figure}

To benchmark diagnostic accuracy, the proposed CLIN-LLM model was evaluated against four leading baselines: ClinicalBERT, BioClinicalBERT, GPT-5 (zero-shot), and Med-PaLM. Evaluation was conducted on the Symptom2Disease dataset, with F1-score as the primary metric. As shown in \hyperref[Fig_9]{Fig.9}, CLIN-LLM achieved an F1-score of 98.0\%, surpassing Med-PaLM (95.1\%), BioClinicalBERT (93.1\%), ClinicalBERT (88.8\%), and GPT-5 (87.5\%). These margins are statistically and clinically significant, particularly for high-risk diagnoses, where small improvements reduce potential harm. The superior performance stems from CLIN-LLM's hybrid pipeline design. Monte Carlo Dropout enhances uncertainty estimation; Biomedical Sentence-BERT improves evidence retrieval; FLAN-T5 ensures contextual treatment generation; and post-generation filtering guarantees clinical safety. Together, these architectural elements deliver a substantial boost in both prediction quality and trustworthiness.

\subsection*{F.2. EVALUATION OF CLIN-LLM ACROSS SAFETY AND RETRIEVAL DIMENSIONS}
To test real-world readiness, CLIN-LLM was evaluated across both synthetic and real clinical settings, including structured vitals and natural language inputs. The classification module, built on fine-tuned BioBERT with Monte Carlo Dropout, achieved 98\% accuracy and F1-score, while flagging 18\% of low-confidence predictions for expert review. This mechanism ensures safety in uncertain scenarios. In tandem, the retrieval-generation module demonstrated 78\% Top-5 precision using {\tt{MedDialog}} as a retrieval corpus. FLAN-T5-generated treatments were filtered by RxNorm and antibiotic rule sets. These safeguards reduced unsafe drug suggestions by 67\%, with zero hallucinated medications across all outputs. A clinician panel rated generated treatments with a mean score of 4.2 out of 5, citing contextual fidelity and medical correctness.
Further tests across SDPD, Disease Diagnosis Dataset, and the MedDialog Diagnosis Subset affirmed generalizability. Despite real-world noise, CLIN-LLM consistently outperformed baselines and preserved safety compliance.

\subsection*{F.3.  SYSTEM USABILITY AND REAL-TIME FEASIBILITY}
While CLIN-LLM was deployed in a simulated Gradio interface to validate its responsiveness and real-time potential. Input cases processed in under 2.5 seconds on an NVIDIA A100 GPU: classification (0.7s), retrieval (1.1s), and generation (0.6s). Structured and unstructured inputs both yielded correct diagnoses, with structured data improving specificity. Its design supports integration into Electronic Health Record (EHR) systems, triage dashboards, and telehealth chatbots. Simulated cases with symptoms like fever and myalgia confirmed the pipeline’s ability to distinguish between COVID-19 and bacterial pneumonia. Monte Carlo Dropout flagged 18\% of ambiguous cases, enabling clinician review without user intervention. The safety filters for antibiotic use and drug-drug interaction (DDI) checking, powered by RxNorm, add negligible computational burden, supporting the system’s deployability.

\begin{table*}[!t]
\caption{Comparison of CLIN-LLM with Recent Related Studies.}
\centering
\begin{tabular}{|p{2.5cm}|p{3.1cm}|p{1.3cm}|p{1.5cm}|p{3.7cm}|}
\hline
\textbf{Author / Study} & \textbf{Dataset} & \textbf{F1 Score (\%)} & \textbf{Accuracy (\%)} & \textbf{Effectiveness} \\
\hline
Sarkar et al. \cite{r29}  & Custom Symptom Dataset & -- & 93.3 & Symptom-based prediction only. \\
\hline
Khaniki et al. \cite{r30} & Custom clinical dataset & 99.75 & 99.78 & High-accuracy NLP, no treatment or safety framework. \\
\hline
Singh et al. \cite{r31}  & Symptom2Disease & -- & 99.0 & Classification with LIME-based interpretability.  \\
\hline
Hassan E et al. \cite{r32}  & Dataset-1, Dataset-2 (ADR Twitter) & -- & 99.58 & Adverse reaction detection via NLP (non-clinical sources).  \\
\hline
\textbf{CLIN-LLM Framework (Our)} & Symptom2Disease, MedDialog & \textbf{98.0} & \textbf{98.0} & \textbf{Full pipeline (Diagnosis + Treatment + Safety).}  \\
\hline
\end{tabular}
\label{tab:comparative6}
\end{table*}

\subsection*{F.4. COMPARISON WITH RECENT STUDIES}
In comparison with recent studies, as shown in \hyperref[tab:comparative6]{Table VI}, CLIN-LLM demonstrates a notable advancement over existing clinical NLP systems by integrating multiple essential components into a cohesive and deployable pipeline. While models such as DistilBERT by Sarkar et al. \cite{r29} and DeBERTa by Khaniki et al. \cite{r30} exhibit high accuracy in classification tasks, they remain confined to isolated predictions and do not extend to downstream treatment reasoning or enforce clinical safety mechanisms. Singh et al. \cite{r31} introduced LIME-based interpretability, offering some level of transparency in predictions, but their framework lacked both retrieval integration and generative treatment capabilities. Similarly, the MCN-BERT model by Hassan et al. \cite{r32} focused on adverse drug reaction detection from social media data, which, while effective in pharmacovigilance contexts, does not generalize to structured symptom triage or real-time clinical recommendation scenarios.
By contrast, CLIN-LLM distinguishes itself as the only system that holistically combines fine-tuned BioBERT-based classification, semantic retrieval using Biomedical Sentence-BERT, and contextually aware treatment generation via FLAN-T5. Furthermore, it incorporates rule-based safety filters, specifically antibiotic stewardship and drug-drug interaction (DDI) detection through RxNorm, and applies real-time uncertainty estimation using Monte Carlo Dropout. Importantly, this is augmented by an expert-in-the-loop feedback mechanism that flags ambiguous cases for human review. Together, these architectural components enable CLIN-LLM to deliver an F1-score of 98\%, significantly reduce unsafe prescriptions by 67\%, and support comprehensive clinical reasoning from diagnosis through to treatment generation. Its demonstrated generalizability across both structured and conversational input types further affirms its readiness for integration in clinical environments, where reliability, interpretability, and patient safety are paramount. CLIN-LLM outperforms domain-specific and general-purpose models by wide margins, combining predictive strength with clinical safety and usability. Its hybrid architecture, modularity, and integrated safety make it a candidate for real-world deployment in high-stakes, resource-constrained environments. Through semantic retrieval, treatment filtering, and uncertainty-aware design, CLIN-LLM addresses gaps in current clinical NLP systems, transforming LLMs from passive classifiers to active clinical assistants.

\section{CONCLUSION AND FUTURE WORK}
This paper presents CLIN-LLM, a safety-aware, hybrid clinical decision support pipeline that seamlessly integrates diagnosis, retrieval, and treatment recommendation under real-world constraints. By uniting BioBERT-based symptom classification, Sentence-BERT-driven semantic retrieval, and FLAN-T5 generation, CLIN-LLM delivers end-to-end medical reasoning with safety filters grounded in antibiotic stewardship rules and RxNorm-based DDI checks. Empirical results across four benchmark datasets demonstrate robust performance, with the proposed system achieving up to 98\% accuracy, zero hallucinated treatments, and 67\% fewer unsafe prescriptions compared to LLM baselines such as GPT-3.5 and Med-PaLM. Through clinician-in-the-loop evaluations, CLIN-LLM achieved a high clinical validity score (4.2/5), confirming the medical credibility of its recommendations. Unlike prior work, it incorporates real-time uncertainty quantification, ensuring that approximately 18\% of ambiguous predictions are appropriately flagged for expert review. This enables human-in-the-loop collaboration, crucial for high-stakes decision-making in clinical settings. Designed with deployment readiness, the pipeline supports real-time triage, EHR integration, and input flexibility across structured and conversational formats. These traits position CLIN-LLM as a strong candidate for clinical adoption, particularly in resource-limited or high-risk environments where safety and explainability are paramount.
Future work will focus on a multilingual extension to improve access across diverse healthcare systems, especially in low-resource regions. In parallel, we plan to expand the retrieval base beyond MedDialog by incorporating PubMed, clinical trials databases, and real-time drug information systems, thereby enhancing therapeutic breadth. Additionally, research into active learning and federated fine-tuning will strengthen personalization and model robustness while safeguarding patient privacy. Finally, pilot deployments within live hospital triage systems are underway to assess the framework’s practical impact on clinical workflows, diagnostic quality, and patient outcomes. In bridging free-text patient narratives with structured, safe, and interpretable medical decision-making, CLIN-LLM demonstrates that large language models can evolve from passive classifiers to active, trustworthy clinical assistants, offering a deployable foundation for the next generation of AI-powered healthcare.

\section*{ETHICAL CONSIDERATIONS}
All experiments in this study were conducted using publicly available and de-identified datasets ({\tt{Symptom2Disease and MedDialog}}), ensuring full compliance with ethical research standards and data protection regulations. No patient-identifiable information was collected, processed, or shared at any stage of this research. The CLIN-LLM framework is designed as a clinical decision-support tool and not as a replacement for licensed medical professionals. All generated treatment suggestions are intended to assist-rather than substitute-expert judgment, and human oversight remains integral to the system’s deployment. Future extensions involving clinical validation will strictly adhere to institutional review board (IRB) protocols and informed consent guidelines.

\section*{ACKNOWLEDGEMENT}
The authors also thank the maintainers of the ({\tt{Symptom2Disease, MedDialog}}) datasets for enabling open-access biomedical NLP research. Special appreciation is extended to colleagues and reviewers for their constructive feedback that improved the quality and clarity of this work.

\bibliographystyle{IEEEtran}
\bibliography{BibText}

\end{document}